\documentclass[runningheads]{llncs}
\usepackage[T1]{fontenc}
\usepackage{graphicx}
\usepackage{subcaption}
\usepackage{placeins}
\usepackage{lipsum}                     
\usepackage{xargs}                      
\usepackage[pdftex,dvipsnames]{xcolor}
\usepackage{verbatim}
\usepackage{amsmath}
\usepackage{makecell}

\begin{document}

\title{Collective Ranking of Environmental Signals through Gaussian Belief Propagation in a Patrolling Robot Swarm}
\titlerunning{Collective Ranking of Environmental Signals by a Patrolling Robot Swarm}
%
\author{Zachary R. Madin\inst{1}\orcidID{0009-0005-0310-4504}\and Connor York\inst{1}\orcidID{0009-0001-8394-2671} \and
Jonathan Lawry\inst{1}\orcidID{0000-0002-3488-5682} \and
Edmund R. Hunt\inst{1}\orcidID{0000-0002-9647-124X}}
\authorrunning{Z. Madin et al.}
%
\institute{University of Bristol, Bristol, UK
\email{\{zachary.madin, connor.york, j.lawry, edmund.hunt\}@bristol.ac.uk}}
\maketitle              
\begin{abstract}
Multi-robot patrolling requires a team to visit all areas of an environment at regular intervals, typically minimising idleness. A practical extension, motivated by security and environmental monitoring, is to additionally form a collective ranking of all patrol locations by some measured signal, a generalisation of the best-of-$n$ problem to the many-option, continuous-valued regime. We observe that the patrol graph admits a natural dual interpretation: it is simultaneously the topology that dictates agent movement and a factor graph over which spatial beliefs can be propagated. Exploiting this equivalence, we apply Gaussian Belief Propagation (GBP), a graph-based algorithm, to collective ranking using unary measurement factors at visited nodes and pairwise smoothness factors along patrol edges. We compare GBP against simple and visit-count-weighted averaging across a range of sensor-noise conditions in simulation, and validate the approach on four Leo Rovers tracking a propagating radio signal in an office lobby. GBP outperforms both baselines on ranking accuracy, mean squared error, and time to consensus. Crucially, as noise increases and the task becomes harder, GBP degrades gracefully in simulation while both averaging methods degrade substantially. Hardware trials reproduce the same performance ordering on a real propagating radio signal, supporting the practical relevance of the simulated results.


\end{abstract}
%
%
%
\section{Introduction}
Surveillance and environmental monitoring applications typically require a large number of areas of interest to be covered and measured repeatedly, and the signals of interest often exhibit spatial structure in which nearby measurements are correlated. Robot swarms are well suited to such problems: they can spread over large regions, pursue multiple objectives in parallel, and cross-verify measurements between agents. A swarm that independently patrols areas of interest while sharing measurements can cover an environment faster than a single agent and be more robust to sensor noise, and by exploiting the spatial structure of the signal it can also produce better estimates at noisy or infrequently visited locations. In this work, we examine how current patrolling methods perform when tasked with monitoring environmental conditions and anomalous signals.
In previous work, we examined how collective consensus formation during patrol can suppress noisy binary anomaly detections, and found that the communication connectivity emergent from a given patrolling algorithm materially affects both the speed and the accuracy of consensus~\cite{madin_collective_2024}. A key open question left by that work is how these findings extend when the quantity being perceived is not a binary anomaly flag but a continuous-valued environmental signal, and when the collective task is not detection but ranking, which we address in this work.

Three information fusion methods of increasing sophistication are compared: a simple mean average of measurements serves as a baseline, followed by a weighted average based on the number of visits to a node, and finally Gaussian Belief Propagation (GBP) which is a graph-based method for distributed inference. Approaches for estimating spatial measurements of environmental variables frequently rely on Gaussian Processes (GPs), which are computationally expensive when the number of observations grows large. 
By contrast we employ GBP, which exploits lightweight local message passing between agents in the swarm and supports incremental node updates to infer environmental values. The graph structure of GBP encodes the spatial relationships between patrol nodes in the graph, allowing beliefs to propagate between adjacent locations.
\begin{figure}
    \centering
    \includegraphics[width=0.75\linewidth]{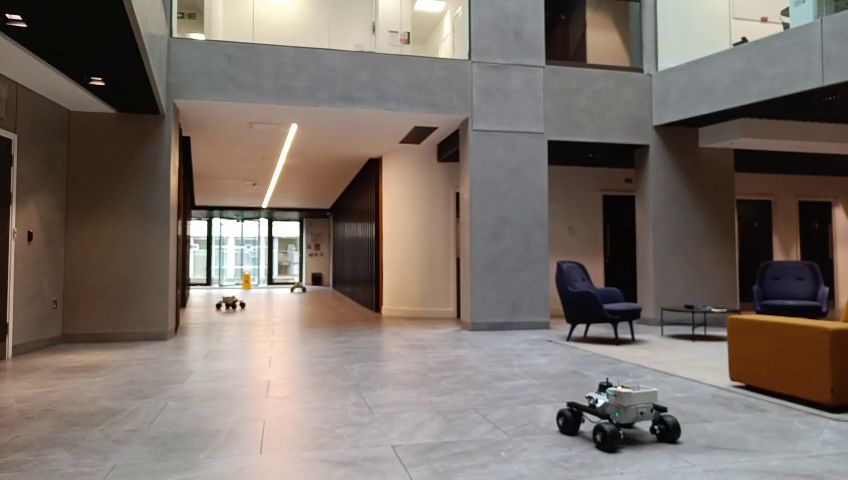}
    \caption{Wheeled rover robots (N=4) autonomously patrolling the lobby of an office building while taking radio measurements along a patrol graph during real trials.}
    \label{fig:real_trial_snapshot}

\end{figure}
We examine the behaviours of patrolling algorithms when the agents are tasked with building an accurate severity rank of signals in an environment. A simulated signal environment is used to examine the performance of the swarm under increasing difficulty of problem, as well as different environments. We analyse the performance of the swarms in terms of the time taken to reach a consensus and the accuracy at that given point. To demonstrate that these results hold for a real signal, a number of trials with Leo Rover robots tracking a radio source are performed for a sub-set of the simulation conditions (Figure~\ref{fig:real_trial_snapshot}).

The key contribution of this paper is the application of GBP to the collective ranking variant of the best-of-n problem in a multi-robot patrolling context, and the novel dual-purpose use of the patrol graph. The graph structure of the problem simultaneously defines agent movement topology and serves as the factor graph structure over which spatial beliefs are propagated. This work is then validated on physical robots with real propagating electromagnetic signals. The paper is organised as follows. Section~\ref{section:related_work} discusses the relevant literature, contextualising the work within the field. Section~\ref{section:methods} presents our methodology. Sections~\ref{section:simulation_results} and~\ref{section:real_results} are the results from simulation and real experiments respectively. Section~\ref{section:conclusion} concludes the paper with a summary of our findings and directions for future work.
\section{Related Work}
\label{section:related_work}
\subsection{Patrolling Algorithms}

The task of environmental monitoring for surveillance or security has been extensively studied in the multi-robot literature \cite{huang_survey_2019,portugal_survey_2011,almeida_recent_2004}. Most frequently the multi-agent patrolling problem is decomposed into a task allocation problem on a graph. 
The environment is abstracted into a graph $\mathcal{G}(\mathcal{V}, \mathcal{E})$ with the graph $\mathcal{G}$ consisting of $\mathcal{V}$ vertices and $\mathcal{E}$ edges. The task of monitoring the environment for a multi-agent system is to assign vertices to be visited to each agent while reducing the time since the vertex was last visited. This is defined as minimising the \textit{``idleness''} within the patrolling literature. For the single agent instance, the task becomes a version of the Travelling Salesman Problem --- the agent must visit every node at least once, minimising the time taken and overlap of distance travelled to achieve an optimal solution. This problem is known to be NP-hard in both the single and multi-agent case \cite{papadimitriou_euclidean_1977,pasqualetti_optimal_2010}, with improvements only recently being made on approximations \cite{karlin_slightly_2021}. 

In this work we employ two algorithms that are extensively used in the literature, chosen for their performance and comparability. Cyclic algorithm for Generic Graphs (CGG) is a method for computing Hamiltonian cycles when the patrol graph allows, and long paths to visit every node in as few steps as possible~\cite{chevaleyre_theoretical_2004}. State Exchange Bayesian Strategy (SEBS) performs a form of Bayesian learning to estimate what the expected idleness of a node is, and determines which node to visit based on a utility function~\cite{portugal_distributed_2013}. This algorithm enables agents to make local and independent decisions while sharing information of visit intentions with agents in their vicinity and is tolerant to communication limitations \cite{yan_multirobot_2016}. The choice of CGG and SEBS in this work is also informed by our prior comparison of ten multi-robot patrolling algorithms on the same ``Cumberland'' patrol graph (Figure~\ref{fig:lobby_graph_map}a), in which SEBS emerged as a strong all-round performer in terms of both idleness minimisation and collective anomaly perception accuracy, while CGG produced a higher-connectivity emergent communication network that favoured rapid consensus at the expense of false-positive suppression~\cite{madin_collective_2024}. Selecting these two algorithms therefore provides a principled contrast between a Bayesian utility-driven patrolling strategy and a fixed-cycle strategy whose behavioural differences are already well-characterised within our experimental framework.

In this work, the patrol graph serves a dual purpose: in addition to defining the topology of agent movement, it defines the discrete set of locations at which environmental measurements are taken, making patrolling the mechanism by which spatial observations are gathered for environmental modelling.

\subsection{Environment Modelling}
The task of monitoring a spatial environmental field using a multi-robot or swarm system has been applied across a wide range of application domains including ocean temperature mapping, pollution monitoring, agricultural sensing, and wildfire detection~\cite{dunbabin_robots_2012,mansfield_survey_2024}. A common and principled approach to modelling a continuous spatial field from discrete robot measurements is Gaussian process (GP) regression~\cite{rasmussen_gaussian_2006}, which provides a probabilistic framework for interpolating between sampled locations and quantifying prediction uncertainty. GP-based methods have been applied to multi-robot systems for mapping scalar environmental fields such as sea-surface temperature, chemical gradients, and water quality parameters~\cite{lin_distributed_2020,xu_spatial_2012,mishra_multiusv_2021}, and offer the advantage of naturally encoding spatial correlation structure into field estimates. However, a well-known limitation of standard GP regression is its computational complexity: inference requires the inversion of the full covariance matrix over all collected observations, an operation that scales as $\mathcal{O}(n^3)$ in the number of data points and $\mathcal{O}(n^2)$ in storage~\cite{rasmussen_gaussian_2006,ambikasaran_fast_2015}, making it increasingly intractable as datasets grow. This cost is particularly problematic in a distributed multi-robot context, where no single agent maintains access to all observations and performing full covariance matrix inversion in a decentralised manner is a non-trivial challenge~\cite{xu_spatial_2012,nabarro_distributed_2026}. Several approximation strategies have been proposed to address these limitations~\cite{mishra_multiusv_2021,viseras_decentralized_2016}, each offering trade-offs between tractability and modelling fidelity.

In settings where the primary objective is not continuous field reconstruction but rather the accumulation of point-value knowledge at specific locations in the environment as is naturally the case in the multi-robot patrolling problem a graph-structured representation of the environment offers a more appropriate and computationally lightweight abstraction. Here, the patrol graph itself defines the discrete set of nodes at which measurements are taken, and rather than fitting a global spatial model, agents can maintain and update local estimates at each node independently as it is revisited over time. A natural and tractable approach in this setting is to combine repeated measurements through a simple running average, accumulating observations across patrol cycles and refining node-level estimates as the patrol progresses~\cite{crosscombe_impact_2022,albani_monitoring_2017}. This approach has an analogue in the broader swarm and multi-robot exploration literature, where agents exploring an environment have been shown to build useful spatial representations through local averaging and consensus-based aggregation of sensor readings acquired during traversal~\cite{jones_distributed_2025,albani_monitoring_2017}. While such averaging methods lack the spatial interpolation properties and principled uncertainty quantification of full GP models, their simplicity and compatibility with distributed, memory-constrained robotic systems make them well-suited to the patrolling context, where the patrol graph structure already imposes a natural discretisation of the environment and the regular re-visitation of nodes provides repeated opportunities to refine estimates over time.

\subsection{Best-of-N and Collective Ranking}
The best-of-n problem, as formalised by Valentini et al. \cite{valentini_bestofn_2017}, requires a swarm of agents to reach collective consensus on which of n available options best satisfies the needs of the collective. In the collective perception literature, this problem has been most commonly instantiated as a binary discrimination task, where agents must determine which of two colours (typically black or white) constitutes the majority of a tiled floor environment. In this abstraction, the quality of each option is directly represented by the proportion of the arena surface covered by a given colour, with the ratio of black to white tiles representing the ground truth that the collective must perceive. While this formulation offers a tractable and well-studied benchmark for evaluating consensus formation strategies, it has limited applicability to real-world deployment scenarios. Security patrol environments, for example, are unlikely to present agents with neatly partitioned binary perceptual landscapes; instead, anomalies of interest may be spatially localised, intermittent, or confounded by environmental noise in ways that a uniform tile ratio does not capture.

Recent work has sought to address this limitation by introducing more principled and varied approaches to the construction of environmental patterns used in collective perception tasks. Benchmarking work by \cite{bartashevich_benchmarking_2019} proposed new task difficulty metrics to better characterise the challenge posed to a collective decision-making system, moving beyond simple majority ratios to consider the spatial arrangement and clustering of features in the environment. Their approach recognises that two environments with identical black-to-white ratios may present very different perceptual challenges depending on how features are distributed a densely clustered arrangement, for instance, may lead to strongly biased local observations for individual agents, increasing the variance of individual measurements and slowing consensus formation. This approach to the formulation of a \textit{hardness} of the best-of-n problem however, still relies on a binary option and has limited applicability to real world ranking of options. In this work, we are interested in a strictly more general task: the collective ranking of all patrol nodes by estimated signal severity, rather than the identification of a single best option. 

This distinction has been formalised by Crosscombe and Lawry~\cite{crosscombe_collective_2021}, who extend the best-of-$n$ framework to enable agents to learn a complete preference ordering over all $n$ options. Shan and Mostaghim~\cite{shan_noiseresistant_2022} directly address this collective ranking task in a spatially distributed robot swarm, comparing a Borda-count ranked voting strategy against a belief-fusion baseline across sweeps of sensor noise, evidence rate, and swarm size. They find that ranked voting is more noise-resistant and scales more favourably with swarm size than belief fusion, at the cost of longer convergence and higher scatter in the swarm's final opinion distribution. They attribute this trade-off to the strong positive-feedback dynamics of fusion-based consensus, which can lock a swarm onto an incorrect option before sufficient evidence has been accumulated: an effect that is exacerbated under high noise or in large swarms. Their three-metric evaluation framework of accuracy, consensus uniformity, and convergence time motivates our use of MSE, Spearman rank correlation, and time-to-consensus as performance metrics in Section~\ref{section:simulation_results} and~\ref{section:real_results}.

A further line of work by Shan and Mostaghim~\cite{shan_manyoption_2024} considers the many-option regime in which the number of discrete options exceeds the swarm size: the relevant regime for our setting, where $40$ patrol nodes are collectively ranked by a swarm of $8$ agents. They compare discrete consensus strategies (ranked voting and distributed Bayesian belief sharing) against a continuous Linear Consensus Protocol (LCP) baseline, and show that the hardness of the consensus task depends on whether option qualities are correlated. When options are weakly correlated, discrete strategies are accurate but slow; when option qualities are strongly correlated or spatially concentrated, fusion mechanisms that treat options as independent become more vulnerable to premature convergence on an incorrect consensus. This distinction is directly relevant to the present work: the radially structured signal field generated by a single electromagnetic emitter produces a spatially correlated option-quality structure that a fusion method exploiting graph topology (such as GBP) should be able to leverage.

\subsection{Averaging Measurements}
In this work, we compare GBP against maintaining a running estimate of the environmental signal at each patrol graph node using two averaging methods. In the first, a simple mean is computed across all measurements recorded at a given node. In the second, a weighted mean is used where the contribution of each robot's local estimate is weighted by the number of visits that robot has made to that node. This results in agents with more numerous visits to a given location exerting proportionally more influence on the fused estimate. Given $R$ robots each holding a local estimate $\hat{z}_{i}^{r}$ at node $i$ with associated visit count $v_{i}^{r}$, the resultant weighted mean estimate is given by:
\begin{equation}
    \hat{z}_{i} = \frac{\sum_{r} v_{i}^{r}\, \hat{z}_{i}^{r}}{\sum_{r} v_{i}^{r}}
    \label{eq:weighted_mean}
\end{equation}
The simple mean averages each robot's running estimate with equal weight, $\hat{z}_{i} = \frac{1}{R}\sum_{r} \hat{z}_{i}^{r}$, and so corresponds to Eq.~\ref{eq:weighted_mean} with $v_{i}^{r} = 1$ for all $r$.

The weighted and simple averaging baselines employed here can be viewed as discrete-graph instantiations of linear consensus~\cite{olfati-saber_consensus_2007}, and serve a similar role to the LCP baseline used by Shan and Mostaghim~\cite{shan_manyoption_2024} in their comparison of discrete and continuous consensus strategies. Both averaging methods aggregate observations node-by-node without reference to the graph's edge structure, and therefore provide a natural baseline against which to isolate the contribution of exploiting the patrol-graph topology during fusion.

\subsection{Gaussian Belief Propagation}

Gaussian Belief Propagation (GBP) is a message-passing algorithm for performing distributed inference over a factor graph, where the goal is to compute the marginal distribution of each variable given a set of observed measurements~\cite{pearl_reverend_1982,loeliger_introduction_2004}. In the context of robotics and multi-agent systems, GBP is particularly attractive because it operates entirely through local computation and nearest-neighbour message passing, requiring no centralised processing or global knowledge of the graph structure~\cite{davison_futuremapping_2019}. More recently, GBP has attracted interest in the context of distributed multi-robot systems, where its local message-passing structure aligns well with the communication constraints of robot swarms~\cite{patwardhan_distributing_2023,jones_distributed_2025}. Each node in the factor graph maintains a Gaussian belief over its associated variable, which is iteratively refined by receiving messages from its neighbours without requiring complete environment information. This makes GBP a natural fit for distributed robotic systems where communication is constrained to local interactions.

A factor graph encodes a joint distribution over unknown variables as a product of local factors, with \textit{variable nodes} representing quantities to be estimated and \textit{factor nodes} encoding probabilistic constraints or measurements~\cite{loeliger_introduction_2004}. Restricting to linear Gaussian factors ensures that all messages remain Gaussian throughout inference and can be computed in closed form~\cite{ortiz_visual_2021}. In the patrol graph setting, variable nodes correspond to patrol graph nodes, each representing the estimated environmental quantity at that location. Unary measurement factors are added as robots visit nodes, progressively tightening each node's belief over time. In this work, the factor graph is initialised with uninformative priors (zero mean, zero precision) at each node and pairwise smoothness factors between patrol nodes, with precisions set as the inverse of edge weight, where edge weight is the traversal distance in metres. Each visit then contributes an additional unary measurement factor via canonical-form addition.

Inference proceeds via the standard sum-product algorithm~\cite{loeliger_introduction_2004}, with messages represented in canonical (precision) form as precision vector $\boldsymbol{\eta}$ and precision matrix $\boldsymbol{\Lambda}$, related to the mean and covariance by $\boldsymbol{\Lambda} = \boldsymbol{\Sigma}^{-1}$ and $\boldsymbol{\eta} = \boldsymbol{\Lambda}\boldsymbol{\mu}$. In this form, the belief at each variable node reduces to a sum of incoming precision vectors and matrices~\cite{ortiz_visual_2021}, with the mean estimate recovered as $\boldsymbol{\mu}_{b_i} = \boldsymbol{\Lambda}_{b_i}^{-1} \boldsymbol{\eta}_{b_i}$.

In a tree-structured factor graph, a single forward-backward pass yields exact inference~\cite{pearl_reverend_1982}. However, patrol graphs are not trees and contain cycles, giving rise to Loopy Belief Propagation (LBP), in which messages circulate iteratively and convergence is no longer guaranteed in general~\cite{weiss_correctness_2000,murphy_loopy_1999}. In practice, LBP has been shown to converge and produce accurate marginals across a wide range of applications~\cite{murphy_loopy_1999}. To promote stable convergence in loopy graphs, \textit{damping} is applied to the message update rule, whereby each new message is computed as a weighted combination of the freshly computed message and the previous message at that edge:
\begin{equation}
    \mu^{(t+1)} = (1 - \alpha)\, \mu^{(t+1)}_{\text{new}}
    + \alpha\, \mu^{(t)},
    \label{eq:damping}
\end{equation}
where $\alpha \in [0, 1)$ is the damping factor and $\mu^{(t)}$ denotes the message at iteration $t$~\cite{ortiz_visual_2021}. Damping reduces the rate at which information propagates around cycles, preventing the runaway reinforcement that can cause divergence in graphs with strong cyclic topologies. In this work, the damping factor ($\alpha = 0.25$) and the number of GBP iterations across the graph per new measurement (fixed at 3) were both determined via a small preliminary sweep, selecting values that reliably produced convergence without introducing oscillation.
\section{Methods}
\label{section:methods}

\subsection{Simulation Environment Initialisation}
\begin{figure}
    \centering
    \includegraphics[width=0.75\linewidth]{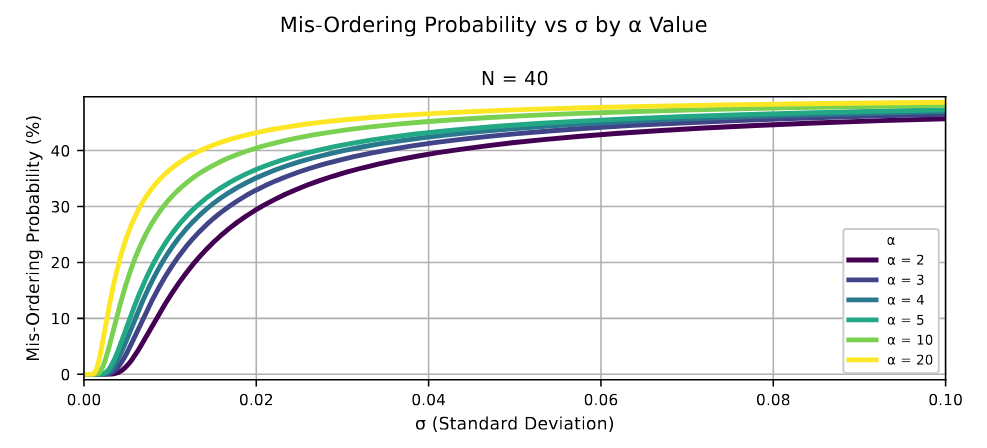}
    \caption{Monte Carlo simulation of mis-ordering probability using Bayes error for different standard deviations of normal distributions used in node means}
    \label{fig:mis-ordering_probability}
\end{figure}

This work exists in the security context of detecting concealed radio frequency emitters in a patrolled environment. We assume the presence of a sensor that returns a \textit{severity} value representing the strength of a detected electromagnetic signal, bounded within the interval $[0, 1]$. Node severity values are drawn from a Beta distribution $\text{Beta}(\alpha, \beta)$, which is well-suited to this purpose given its support over the unit interval and its flexibility in representing a wide range of unimodal and skewed distributions through the choice of shape parameters. An initial examination of multiple $\alpha = \beta$ values was performed as seen in Figure \ref{fig:mis-ordering_probability}, with the final value being $\alpha=\beta=5$. Each node in the patrol graph is assigned one of these sampled severity values, providing a continuous-valued analogue to the discrete option qualities typically considered in the best-of-n literature~\cite{valentini_bestofn_2017}, where the number of distinguishable options is small and their qualities are clearly separated. Here, we instead examine the best-of-n problem over a continuous interval, where the difficulty of correctly ordering node intensities depends on the degree of overlap between their underlying distributions.

To quantify the difficulty of the classification task, we characterise the \textit{mis-ordering probability} the probability that a noisy sensor reading causes an agent to incorrectly order a pair of node values as a function of the standard deviation $\sigma$ of the sensing noise. Within our simulation for simplicity we assume the noise of a measured signal is Gaussian. For a given $\sigma$, we model each node's observed severity as a Gaussian distribution centred on its true Beta-sampled value with standard deviation $\sigma$, and compute the pairwise Bayes error between all node distributions. A Monte Carlo simulation is performed across a range of $\sigma$ values to produce an empirical mapping from noise level to expected mis-ordering probability, providing a principled basis for selecting experimental noise conditions that span a meaningful range of task difficulty. The noise levels used in our experiments, expressed as mis-ordering probabilities ranging from $0\%$ to $45\%$ in increments of $5\%$, are selected according to this mapping and reported in Table~\ref{tab:sim_params}.

To mirror the realistic spatial structure of an electromagnetic emitter, the Beta-sampled severity values are assigned to patrol graph nodes according to their radial distance from a simulated signal centre, such that nodes closer to the centre receive higher severity values. This introduces a spatial correlation into the node value distribution that reflects the physically motivated assumption that signal strength decays with distance from the source, and provides a structured environmental field that can be exploited by fusion algorithms that account for spatial relationships between nodes. It is worth noting however that the physical experiment is subject to multipath effects of the radio signal, and will not follow the same smoothness behaviour as the simulation.

\subsection{Simulator}
The simulator used in this paper is written in Python, and is a discrete 2D grid world where agents are represented by occupying a single cell. They can move to all eight surrounding cells, and avoid colliding with walls and other agents. Each patrolling algorithm is implemented directly as well as each of the fusion methods. At time zero, agents determine which node to visit on their patrol, and start navigating to it synchronously. Once an agent arrives at a node a measurement is taken from the node's normal distribution, which is then incorporated to the fusion method. Communication happens inter-agent without a central controller and parameters such as the communication frequency and range can be varied. The simulator uses existing maps of varying size and graph dimension that are frequently used in the literature \cite{portugal_performance_2011}.

\begin{figure}[htbp]
  \centering

  \begin{subfigure}[b]{0.55\textwidth}
    \centering
    \includegraphics[width=\textwidth]{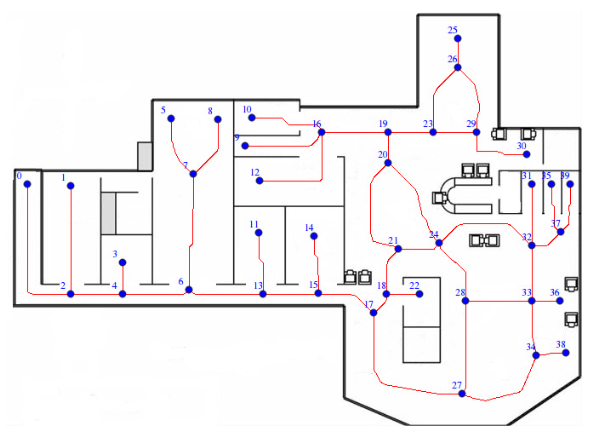}
    \caption*{(a)}
  \end{subfigure}
  \hfill
  \begin{subfigure}[b]{0.35\textwidth}
    \centering
    \includegraphics[width=\textwidth]{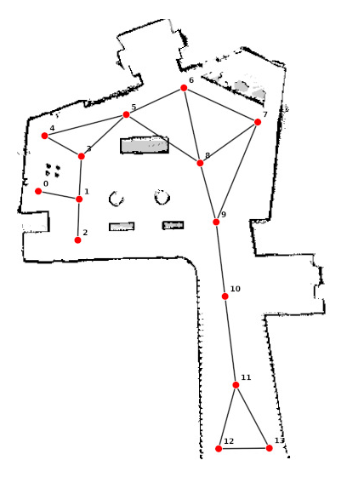}
    \caption*{(b)}
  \end{subfigure}

    \caption{Patrol graph environments used in experiments: (a) ``Cumberland'' (40 nodes, simulation trials) and (b) Bristol Lobby (14 nodes, real-robot trials).}
  \label{fig:lobby_graph_map}
\end{figure}
\section{Simulation Results}
\label{section:simulation_results}
This section will discuss the results gained from the simulation experiments. A brief note will be made on the communication range and the influence it has on convergence and performance. The same algorithms and fusion methods were tested on two different maps with different topology: one in simulation (this section) and one real-world map (Section~\ref{section:real_results}). We examined the performance of each combination of patrolling algorithm and fusion method 20 times. The parameters for the simulation are shown in Table~\ref{tab:sim_params}.

Consensus convergence is assessed by computing the standard deviation $\sigma_{inter}$ of node value estimates across all agents at each timestep. When $\sigma_{inter} < \varepsilon$ for all nodes in the patrol graph, the system is considered to have reached consensus, indicating that inter-agent estimates have stabilised and are no longer varying significantly across the robot population. 
Communication range was found empirically to have no substantial effect on fusion performance beyond a minimum threshold required for consensus; global communication was therefore adopted across all experiments to isolate the effects of patrolling algorithm and fusion method, as different patrolling algorithms exhibit different minimum communication requirements that would otherwise hinder direct comparison between algorithms.
Studying the interaction between fusion method and emergent connectivity in the ranking setting is left for future work.

\begin{table}[ht]
    \centering
    \caption{Simulation parameter values used across all experiments.}
    \begin{tabular}{ll}
        \hline
        \textbf{Parameter} & \textbf{Value} \\
        \hline
        Number of robots & 8  \\
        Communication range & Global \\
        Communication timeout method & Upon new data \\
        Duration of experiment & 10,000\,steps or consensus \\
        Patrol graph nodes & 40 \\
        Mis-ordering probability & 0 - 45\% \\
        Simulations per algorithm  & 20 \\
        Value of $\alpha = \beta$ & 5 \\
        GBP iteration count & 3 \\
        Damping Factor $\alpha$ & 0.25 \\
        
        \hline
    \end{tabular}
    \label{tab:sim_params}
\end{table}

\subsection{Performance Metrics}
Performance is evaluated across three metrics. Convergence time is recorded as the number of simulation timesteps until the standard deviation of estimates across all agents falls below an empirically determined threshold $\varepsilon$ for every node in the patrol graph, indicating that the swarm has reached a stable consensus. Once consensus has been reached, the following metrics determine performance of the swarm. Summed mean squared error is computed by summing the squared difference between each agent's estimate and the ground truth value across all agents and nodes. Spearman's rank correlation coefficient $\rho$ measures the distance between the swarm's collective node ordering and the ground truth ranking.

\subsection{Results}
\
\begin{figure}
    \centering
    \includegraphics[width=\linewidth]{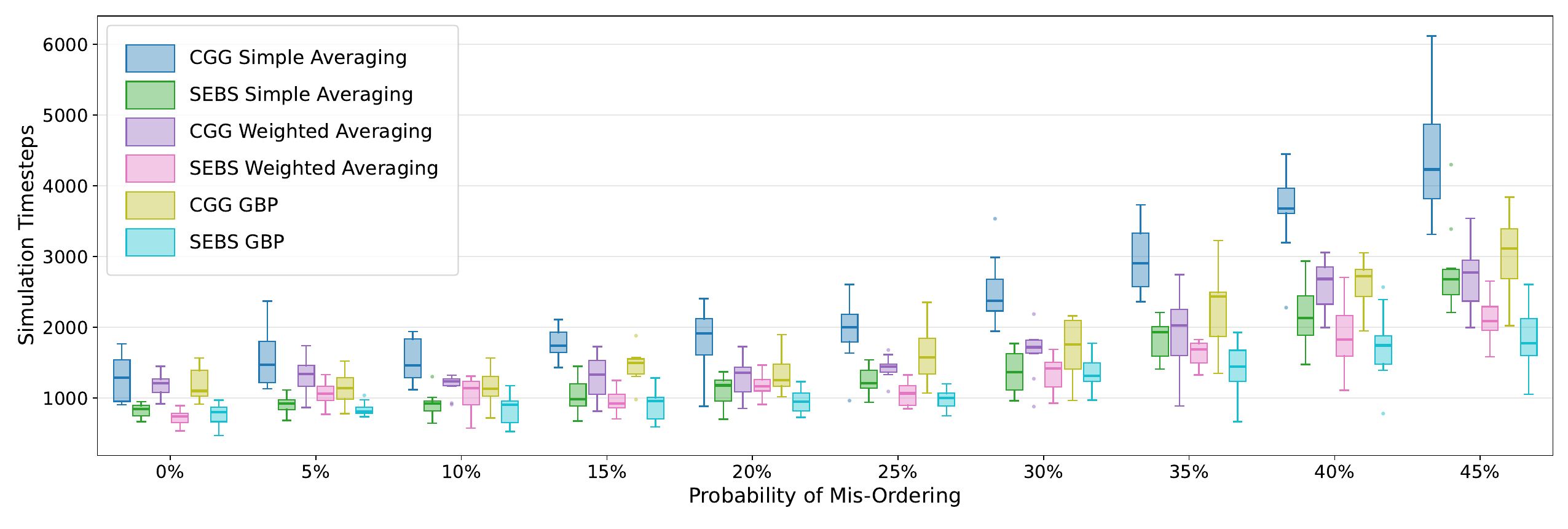}
    \includegraphics[width=\linewidth]{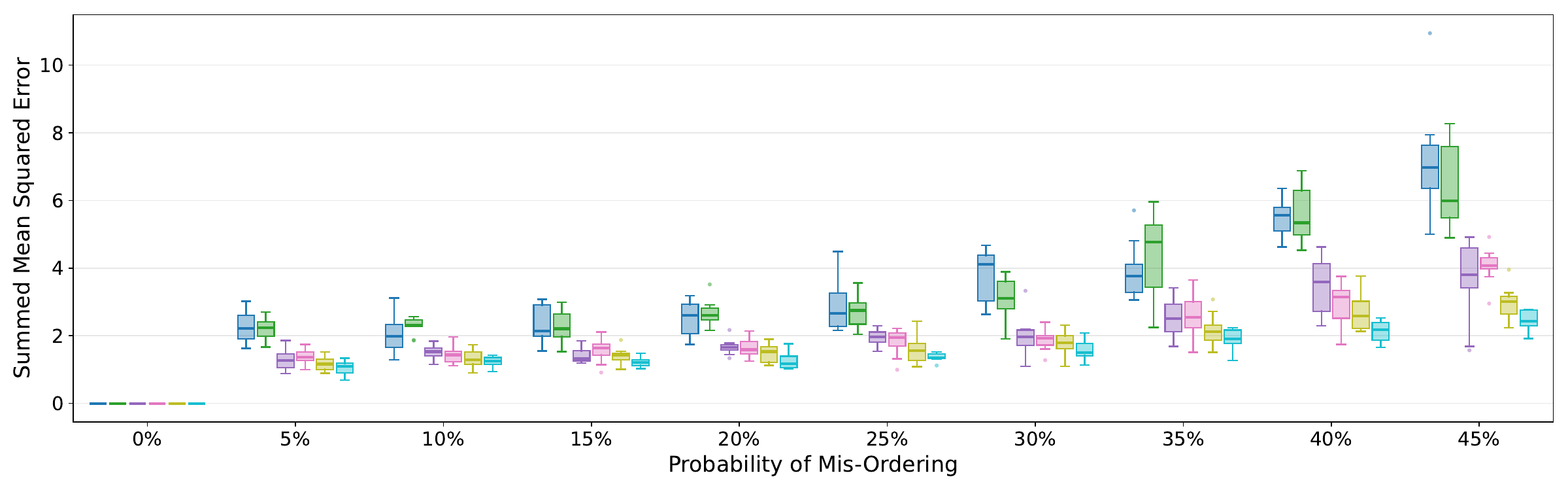}
    \includegraphics[width=\linewidth]{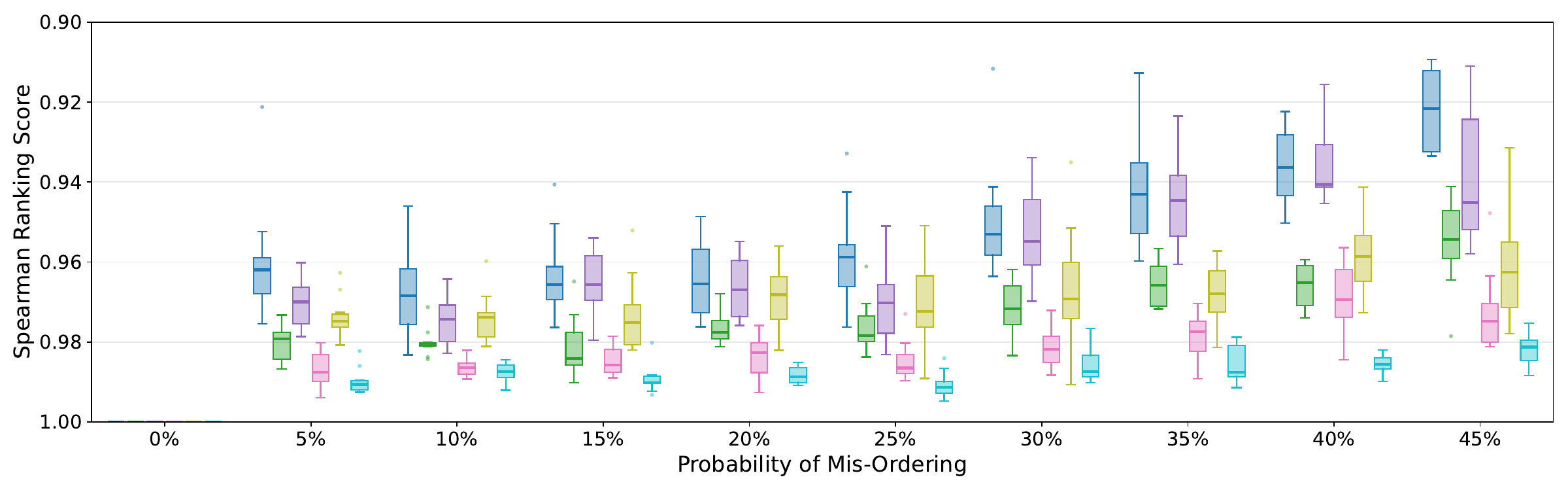}

    \caption{Performance metrics by patrol algorithm and fusion method, for varying mis-ordering probabilities with a simulated 8 robot swarm (20 runs per condition). Top: Consensus time. Middle: Mean Squared Error summed across the simulated swarm. Bottom: Spearman Ranking Score (axis inverted so that lower indicates better performance).}
    \label{fig:sim_results_mse}

\end{figure}

Across all conditions, the three fusion methods exhibit a clear performance hierarchy (Figure~\ref{fig:sim_results_mse}). Simple averaging performs worst across all three metrics, with both MSE and Spearman rank distance degrading substantially as mis-ordering probability increases beyond 20\%. Weighted averaging consistently outperforms simple averaging, achieving lower MSE and better ranking accuracy with a reduced time to consensus, reflecting the benefit of weighting agent estimates by visit count. GBP outperforms both averaging methods across all metrics at all noise levels.
Of the two patrolling algorithms, SEBS outperforms CGG across both performance metrics and reaches consensus in fewer timesteps, suggesting that its Bayesian utility-driven node selection produces more informative measurement sequences than the fixed cyclic traversal of CGG.
Notably, GBP demonstrates a robustness to increasing noise that the averaging methods do not. As mis-ordering probability increases beyond 25\%, both averaging methods show marked degradation in MSE and Spearman ranking score, while GBP maintains accurate performance with minimal degradation even at the highest noise conditions tested. This is attributable to GBP's exploitation of spatial correlations encoded in the patrol graph topology, which allows beliefs at noisy or infrequently visited nodes to be informed by estimates at neighbouring nodes, an advantage unavailable to methods that treat each node's measurements independently.
\section{Real Robot Experiments}
\label{section:real_results}

\subsection{Radio Signal Used}
The conclusions from the simulated results were examined in a real world environment, with the agents sampling an environment with a propagating radio signal (Figure~\ref{fig:real_trial_snapshot}). In these experiments, a simple measure of signal intensity (RSSI) is used to represent the power of the signal received by the agents. Each agent performs their role of patrolling, while taking measurements when arriving at a node. The measurement values are fed into the three fusion methods examined in the simulation results, and compared for the Mean Squared Error, Spearman Rank distance and time to convergence. As the behaviour of the swarm is not dictated by the measurements but by the patrolling algorithm, all three methods are computed for the same trial data and then plotted for comparison. Ten trials of the patrolling algorithm SEBS were performed in the environment.

For the signal to be measured, an XBee S2C Zigbee~\cite{noauthor_xbeexbee-pro_nodate} is used with a single static device acting as the transmitter, and each agent having a device as a receiver. The XBee Zigbee device operates on the 2.4 GHz frequency, and transmits with a power of 3.1 mW (+5 dBm). Signal packets are transmitted at 20 Hz, and the receiver sensitivity is $-100$ dBm. This device and method were selected to satisfy the size and power constraints of the real robotic platform, while the moderately low transmit power ensures a detectable decay in signal strength across the environment, producing the desired signal propagation pattern. The ground truth of the signal prior to experiment for ranking purposes was determined by taking repeated recordings over time and averaging across values.

The environment in which the robots patrol and sample the radio signal is an office lobby covering an area of $20\times14$ m. Nodes on the graph structure are distributed such that the distance between neighbouring nodes is no greater than 5 metres to ensure fair coverage of the environment. A representation of the navigation graph and mapped environment can be seen in Figure~\ref{fig:lobby_graph_map}(b).

A small swarm of homogeneous robotic agents is used in this patrolling trial, each consisting of a non-holonomic rover-style platform. Four Leo Rovers are used in this experiment, each equipped with a battery pack, LiDAR, XBee S2C Zigbee transceiver, Latte Panda SBC and Raspberry Pi, performing combined high-level SLAM and navigation.

\begin{figure}
    \includegraphics[width=\textwidth]{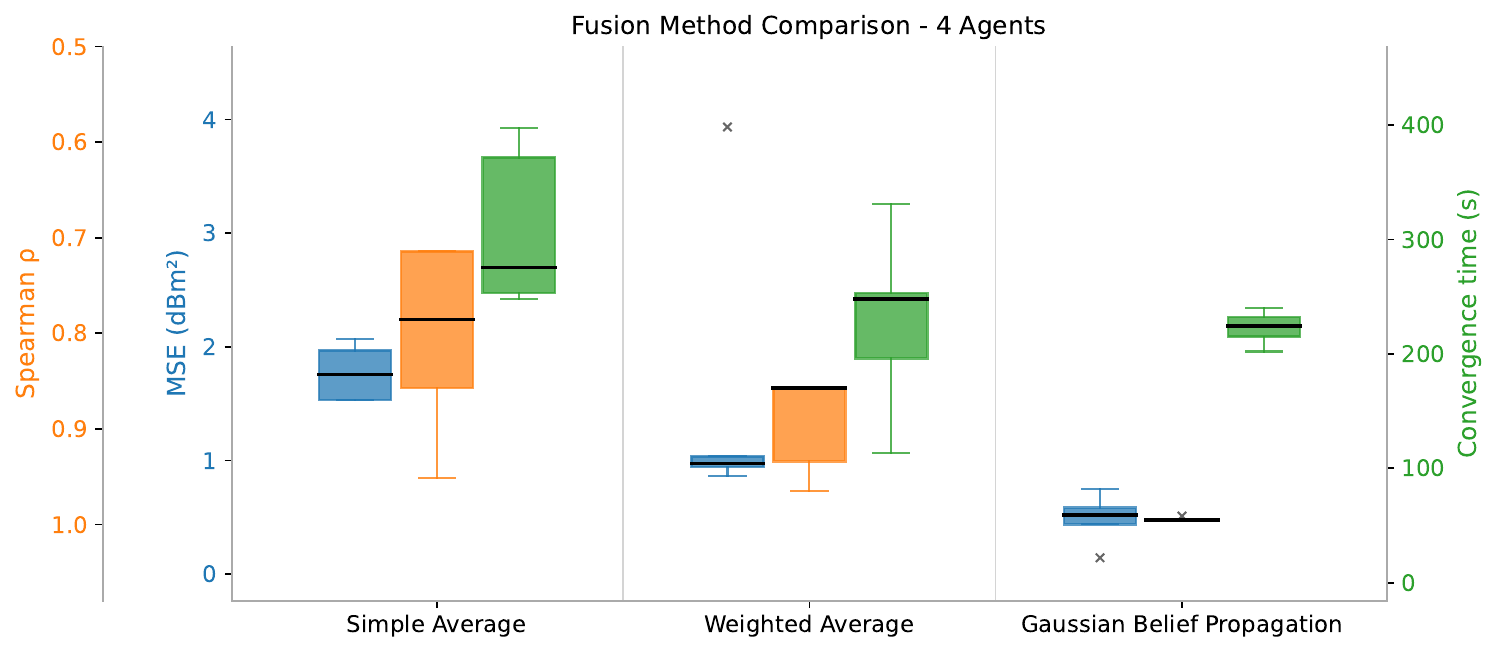}
    \caption{Comparison between fusion methods used in real trial data with only SEBS patrolling algorithm. Mean Squared Error, Spearman Ranking $\rho$ on left vertical axis (inverted), and convergence time on right vertical axis. Each fusion method is computed from the same 10 experimental runs.}
    \label{fig:real_results}
\end{figure}

Simulation operates on normalised severity values in [0,1] while real trials measure RSSI in dBm, so absolute MSE magnitudes are not directly comparable across the two settings; only the relative ordering of fusion methods is. As shown in Figure~\ref{fig:real_results}, the hierarchy observed in simulation is preserved: simple averaging performs worst, weighted averaging improves on it, and GBP achieves the lowest MSE, highest Spearman ranking score, and shortest median convergence time. This holds despite real signal propagation being subject to multipath interference, environmental occlusion, and hardware noise not present in simulation. 
\FloatBarrier

\section{Conclusion \& Future Work}
\label{section:conclusion}
We examined a dual-objective robot swarm performing patrolling and environmental monitoring in a security context, framing the task as a collective ranking problem over continuous-valued signals at a large number of patrol nodes. Using two established idleness-minimisation algorithms, we compared Gaussian Belief Propagation against simple and weighted averaging as fusion methods, and found that GBP yields more accurate ranking and faster convergence at the cost of additional computational requirements of the platform. The same performance hierarchy was observed when deploying the system on real robots tracking a propagating radio signal, establishing GBP as a reliable distributed method for collective environmental monitoring. These results are directly actionable for security applications in which a human operator, with limited time to verify a swarm's output, must prioritise the most severe readings.

Our findings align with and extend observations from the discrete collective decision-making literature. Shan and Mostaghim~\cite{shan_noiseresistant_2022,shan_manyoption_2024} argue that fusion strategies which enforce consensus through positive feedback can lock a swarm onto an incorrect collective decision when measurements are noisy or option qualities are spatially correlated. This is due to treating each option independently and reinforcing shared beliefs among nearby agents. The averaging methods considered here show a related pattern as measurement noise increases: by treating each patrol node as an independent estimation problem, they degrade markedly under high noise even though the underlying signal field is spatially correlated. GBP by contrast, replaces pure consensus enforcement with a structural constraint: the patrol graph acts as a factor graph that propagates beliefs between adjacent variable nodes, so a noisy or infrequently visited node inherits information from its better-sampled neighbours, allowing the swarm to maintain a globally consistent estimate.

A natural direction for future work is to incorporate agent pose uncertainty into the fusion framework. This problem formulation lends itself well to GBP, where the uncertainty of measurements is added as an additional two dimensions to the inference framework. This addition leaves the distributed message passing benefits unchanged, while naturally propagating pose uncertainty into node estimates. This would enable equivalent environmental monitoring on platforms with less accurate localisation, in turn making larger swarm sizes tractable.

%
%
\begin{credits}
\subsubsection{\ackname}ZRM is supported by a University of Bristol PhD Scholarship. CY is supported by UK FCDO Services. ERH is supported by the Royal Academy of Engineering under the Research Fellowship programme. Simulations were performed on University of Bristol Self-Service Cloud.


\end{credits}
%
%
%
\bibliographystyle{splncs04}
\typeout{}
\bibliography{references_edit}

\end{document}